%% file: neurips_2020.tex
\documentclass{article}

    \PassOptionsToPackage{numbers, compress}{natbib}

    \usepackage[preprint]{neurips_2020}

\usepackage[utf8]{inputenc} 
\usepackage[T1]{fontenc}    
\usepackage{hyperref}       
\usepackage{url}            
\usepackage{booktabs}       
\usepackage{amsfonts}       
\usepackage{nicefrac}       
\usepackage{microtype}      
\usepackage{multirow}

\usepackage{graphicx} 
\usepackage{bbm}
\usepackage{subcaption}
\usepackage[title]{appendix}

\usepackage{xcolor}

\title{Assessing Post-Disaster Damage from Satellite Imagery using Semi-Supervised Learning Techniques}

\author{%
  Jihyeon Lee\textsuperscript{1,3}
  \And
  Joseph Z. Xu\textsuperscript{1}
  \And
  Kihyuk Sohn\textsuperscript{1}
  \And   
  Wenhan Lu\textsuperscript{1}
  \And
  David Berthelot\textsuperscript{1}
  \And
  Izzeddin Gur\textsuperscript{1}
  \And
  Pranav Khaitan\textsuperscript{1}
  \And
  Ke-Wei (Fiona) Huang\textsuperscript{2}
  \And
  Kyriacos Koupparis\textsuperscript{2}
  \And
  Bernhard Kowatsch\textsuperscript{2} \\
  \\
  
  \textsuperscript{1}Google Research, \textsuperscript{2}United Nations World Food Programme, \textsuperscript{3}Stanford University\\
  \\
  
  \texttt{jihyeon@cs.stanford.edu},\texttt{\{jzxu,kihyuks,wenhan,dberth,izzeddin\}@google.com}

}

\begin{document}

\maketitle

\begin{abstract}
To respond to disasters such as earthquakes, wildfires, and armed conflicts, humanitarian organizations require accurate and timely data in the form of damage assessments, which indicate what buildings and population centers have been most affected. Recent research combines machine learning with remote sensing to automatically extract such information from satellite imagery, reducing manual labor and turn-around time. 
A major impediment to using machine learning methods in real disaster response scenarios is the difficulty of obtaining a sufficient amount of labeled data to train a model for an unfolding disaster. This paper shows a novel application of semi-supervised learning (SSL) to train models for damage assessment with a minimal amount of labeled data and large amount of unlabeled data. We compare the performance of state-of-the-art SSL methods, including MixMatch \cite{berthelot2019mixmatch} and FixMatch \cite{sohn2020fixmatch}, to a supervised baseline for the 2010 Haiti earthquake, 2017 Santa Rosa wildfire \cite{gupta2019xbd}, and 2016 armed conflict in Syria \cite{Xu2019}. We show how models trained with SSL methods can reach fully supervised performance despite using only a fraction of labeled data and identify areas for further improvements. 
\end{abstract}

\section{Introduction}
\label{intro}
\input{sections/intro}
\section{Related Work}
\label{relatedworks}
\input{sections/relatedworks}

\section{Data}
\label{data}
\input{sections/data}

\section{Approach}
\label{methods}
\input{sections/methods}

\section{Experiments \& Results}
\input{sections/results}
\label{results}

\section{Conclusion}
\input{sections/conclusion}
\label{others}

\begin{ack}
This work is a collaboration between Google Research and the United Nations World Food Programme (WFP) Innovation Accelerator. The WFP Innovation Accelerator identifies, supports and scales high-potential solutions to hunger worldwide. We support WFP innovators and external start-ups and companies through financial support, access to a network of experts and a global field reach. We believe the way forward in the fight against hunger is not necessarily in building grand plans, but identifying and testing solutions in an agile way. The Innovation Accelerator is a space where the world can find out what works and what doesn’t in addressing hunger - a place where we can be bold, and fail as well as succeed.
\end{ack}

\bibliographystyle{plain}
\bibliography{references}

\newpage

\begin{appendices}

\section{Visualization of Satellite Imagery}
\label{sec:supp_data_visualization}

In this work, we consider evaluating on three datasets: Santa Rosa~\cite{gupta2019xbd}, Haiti and Aleppo~\cite{Xu2019}. In this section, we provide visualization of full and cropped satellite imagery for each dataset.

\begin{figure*}[h]
    \centering
    \includegraphics[width=\textwidth]{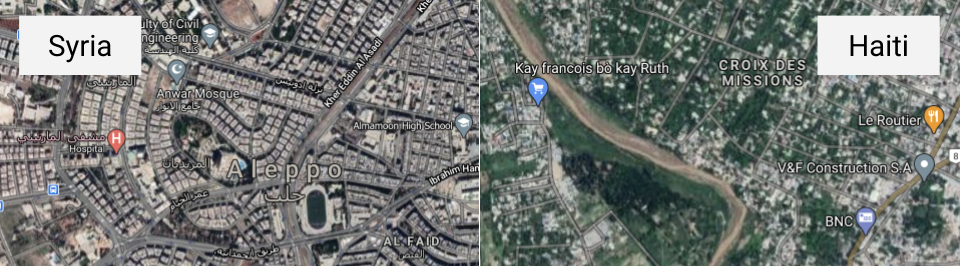}
\caption{Comparison of imagery in two different regions, Syria and Haiti. Building layout, construction material, level of vegetation, appearance of damage (based on disaster type and buildings), and other factors make it difficult for models to generalize to unseen regions.}
\label{fig:differentregion}
\end{figure*}

\begin{figure*}[h]
    \centering
    \includegraphics[width=0.5\textwidth]{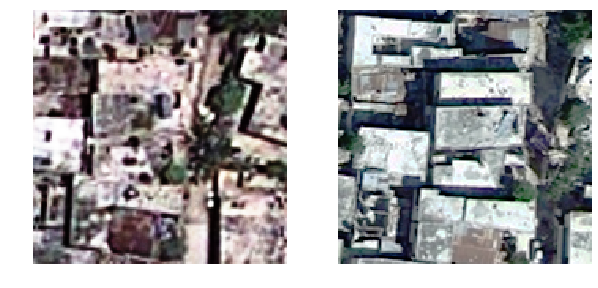} \\
    \includegraphics[width=0.5\textwidth]{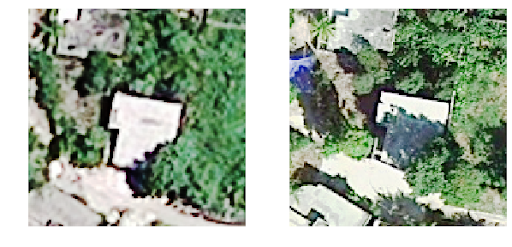} \\
    \includegraphics[width=0.5\textwidth]{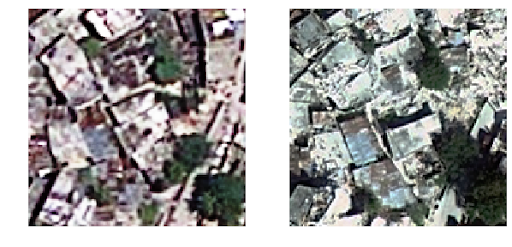}
\caption{Example of images cropped around buildings from Haiti.}
\label{fig:cropped}
\end{figure*}

\section{Details of Experiments}
\label{sec:supp_exp_details}

We provide additional details of experiments for reproducible research.


\subsection{Hyperparameters}
\label{sec:supp_exp_details_hparam}

For the fully supervised and MixMatch experiments, we used a batch size of 64, learning rate of 0.002, and weight decay of 0.002. We used a Beta distribution of $Beta(0.5)$ for mixup. For FixMatch, we used a batch size of 64, learning rate of 0.03, and weight decay of 0.0005. Additionally, the ratio for unlabeled data $\mu = 3$, pseudo-label loss weight $\lambda = 1$, and the confidence threshold $\tau = 0.95$.

For MixMatch and FixMatch experiments, rather than decaying the learning rate, we evaluate models using an exponential moving average of their parameters with a decay rate of 0.999.

\section{Ablation Study on Augmentation Policy}
\label{sec:ablation_augmentation_policy}

We used the RandAugment~\cite{cubuk2019randaugment} augmentation policy for FixMatch, which randomly selects transformations for each sample in a minibatch from a collection of transformations (e.g. color inversion and geometric changes, such as rotation and translation). We also apply Cutout~\cite{devries2017Cutout}, which sets a random square patch of pixels to gray. We conduct an ablation study to examine which operations play a key role in FixMatch, testing 7 settings: 

\begin{enumerate}
    \item Cutout only
    \item RandAugment only with color transformations
    \item RandAugment and Cutout with color transformations
    \item RandAugment only with geometric transformations
    \item RandAugment and Cutout with geometric transformations
    \item RandAugment only with color and geometric transformations
    \item RandAugment and Cutout with color and geometric transformations
\end{enumerate}

To clarify, the setting used in the Results section is RandAugment and Cutout with color and geometric transformations. We use 50 labeled and 50 unlabeled examples in training for all experiments. We run each setting 5 times and report the average and standard deviation of classification performance in Table \ref{tab:ablation}. Generally, Cutout seemed to play a more signficant role than RandAugment, similarly to the finding in the original FixMatch paper where Cutout enabled strong results on CIFAR-10 \cite{sohn2020fixmatch}. Between geometric and color transformations, the former seemed more influential, though the difference becomes marginal when Cutout is applied.

\begin{table}
\centering
\begin{tabular}{c|c|c|c|c|c}
RandAugment Transformation & Cutout                & Accuracy & Haiti & SoCal & Syria \\ \hline
none                          & yes                   & mean     & 0.60  & 0.97  & 0.83  \\
                           &                       & stdev    & 0.06  & 0.01  & 0.03  \\
color                      & no                    & mean     & 0.47  & \textbf{0.98}  & 0.61  \\
                           &                       & stdev    & 0.07  & 0.01  & 0.18  \\
color                      & yes                   & mean     & 0.66  & 0.92  & 0.87  \\
                           &                       & stdev    & 0.06  & 0.02  & 0.02  \\
geo                        & no                    & mean     & 0.56  & 0.97  & 0.73  \\
\multicolumn{1}{l|}{}      & \multicolumn{1}{l|}{} & stdev    & 0.01  & 0.01  & 0.15  \\
geo                        & yes                   & mean     & 0.68  & 0.97  & 0.87  \\
\multicolumn{1}{l|}{}      & \multicolumn{1}{l|}{} & stdev    & 0.05  & 0.01  & 0.02  \\
color + geo                & no                    & mean     & 0.63  & \textbf{0.98}  & 0.82  \\
\multicolumn{1}{l|}{}      & \multicolumn{1}{l|}{} & stdev    & 0.02  & 0.01  & 0.07  \\
color + geo                & yes                   & mean     & \textbf{0.75}  & 0.97  & \textbf{0.89}  \\
\multicolumn{1}{l|}{}      & \multicolumn{1}{l|}{} & stdev    & 0.10  & 0.01  & 0.01 
\end{tabular}

\caption{Ablation study of applying different transformations of RandAugment and Cutout. We report the average and standard deviation over 5 runs per setting.}
\label{tab:ablation}
\end{table}

\end{appendices}

\end{document}

%% file: sections/intro.tex
When a humanitarian crisis such as a natural disaster occurs, crisis responders need to know the locations of affected populations to facilitate relief efforts. The locations and density of damaged buildings serve as a useful proxy to estimate this information \cite{dellacqua}. One approach to identify them is remote sensing: expert analysts compare pre- and post-disaster satellite imagery of the affected region and mark the locations of damaged buildings. However, this is time-consuming (fewer than 100 buildings assessed per hour per person), challenging to scale, and prone to error \cite{Valentijn2020transferability,KerleAccuracy2010}.

Machine learning (ML) has been utilized as an efficient tool to automate the damage assessment process \cite{Cooner2016,Ji2018,Duarte2018,gupta2019xbd,Xu2019,gupta2019xbd,gupta2020rescuenet,weber2020building}. Models are trained to distinguish between images of damaged and undamaged buildings using expert-labeled images of past disasters. The trained models can analyze entire cities in a matter of minutes when deployed in modern data centers. 

It remains challenging to build accurate models for new disasters. ML has traditionally relied on datasets with sufficient variation and coverage so that models generalize to unseen examples at inference time. This is not possible in our domain because there is a limited number of past disasters, and the appearance of each disaster or geographical region varies widely in their layout of buildings, construction material, vegetation, appearance of damage, etc. \cite{nex2019structural} (see Figure \ref{fig:differentregion} in Appendix). Furthermore, even if the model has been pre-trained on the same disaster type and location, there is noise inherent to satellite imagery due to changing landscapes, seasonal variation, cloud cover, and other factors that causes the inference data to systematically differ from the training data. Therefore, models trained only on past disasters will likely under-perform in new disasters.

We can avoid the generalization problem by training models on data from a new disaster after it strikes. Models trained in this way must use only a small number of labeled training examples, since manual expert labeling is time-consuming. 
At the onset of a new disaster, labeled data is limited limited, but an abundant amount of unlabeled satellite imagery can be automatically extracted from the region. Recent advances in semi-supervised learning (SSL) techniques show that algorithms combining labeled and unlabeled training data can achieve performance comparable to fully supervised algorithms trained on orders of magnitude more labeled data \cite{berthelot2019mixmatch,sohn2020fixmatch,Berthelot2020ReMixMatch}. By using SSL techniques, we can train accurate damage assessment models for new disasters without spending much time gathering manual labels. 

In this paper, we apply two SSL techniques, MixMatch \cite{berthelot2019mixmatch} and FixMatch \cite{sohn2020fixmatch}, to train building damage detection models using a limited amount of labeled data. We evaluate on three different disasters, varying the amount of labeled examples. We show that the models can get close to the level of fully supervised results using only a fraction of the labeled data. 

%% file: sections/relatedworks.tex
\paragraph{Machine Learning in Damage Building Assessment}

Past studies from fully supervised settings have successfully applied machine learning approaches to building damage detection from satellite imagery. The public xBD dataset \cite{xView2DatasetCreation,gupta2019xbd} was released alongside the xView2 Challenge \cite{xView2ChallengeWebsite}, providing large-scale satellite imagery, building polygons, and ordinal labels that denote damage level across 19 disasters with the task of per-pixel classification. The first-place approach \cite{xView2ChallengeWinner} had two stages, initially training a localization model with only pre-disaster images and then using the weights to initialize a Siamese Neural Network for building classification that shares weights between the pre-disaster and post-disaster images. Gupta et al. \cite{gupta2020rescuenet} proposed an end-to-end approach, first extracting multi-scale image features, feeding them into a segmentation head to predict buildings independently on the pre- and post-disaster images, and finally classifying each pixel. Weber et al. \cite{weber2020building} trained a single network, modeling the task as semantic segmentation. 

The above methods generate training and validation data from the same distribution, so they do not address the problem of running inference for a newly unfolding disaster with minimal data. Xu et al. \cite{Xu2019} developed models for the Haiti, Mexico, and Indonesia earthquakes and conducted cross-region generalization experiments, showing how models pretrained on past disasters did not perform well in a new region with no or minimal labeled data. We use this method as a baseline in our experiments. Valentjin et al. \cite{Valentijn2020transferability} ran experiments on 13 disasters differing in hazard type, geographical region, and satellite parameters and found performance varied significantly across test disasters, whether or not data from the test disaster was included in training. 

\paragraph{Semi-Supervised Learning Approaches}

Semi-supervised learning (SSL) provides approaches to alleviate the need for large amounts of labeled data by leveraging unlabeled data. There is a class of SSL methods for deep networks that perform pseudo-labeling (or self-training)~\cite{mclachlanSelfTraining1975,Rosenberg2005SelfTraining,Xie2020SelfTraining,scudder1965proboferror}, generating artificial labels from the unlabeled data and involving a minimal amount of human labor \cite{zhang2018mixup,berthelot2019mixmatch,Berthelot2020ReMixMatch,sohn2020fixmatch}. These techniques rely on consistency regularization \cite{Sajjadi2016consistency,Laine2016consistency}, which encourages the model to output predictions of a similar distribution across perturbations of a given input. 
MixMatch \cite{berthelot2019mixmatch} adds other types of regularization, using MixUp \cite{zhang2018mixup} to encourage convex behavior “between” examples by generating weighted combinations of labeled and unlabeled ones. FixMatch \cite{sohn2020fixmatch} presented a simpler approach that achieved state-of-the-art performance on common SSL benchmarks, such as CIFAR \cite{CIFAR} and STL \cite{Coates2011STL}. The methods differ in how they use the pseudo-labels to calculate loss. For a given unlabeled image, MixMatch creates a guess label based on its weakly augmented versions and calculates loss based on how well the model predicts that label. FixMatch also generates a pseudo-label from weak augmentations, but it calculates loss on whether the model is able to predict the label on strongly augmented versions. The driving assumptions of pseudo-labeling SSL methods should hold true for satellite imagery and provide a way to take advantage of unlabeled data.

%% file: sections/data.tex
We evaluated our approach on three disasters.
In addition to the Santa Rosa wildfire \cite{gupta2019xbd}, we generated our own datasets for the 2010 Haiti earthquake and 2016 snapshot of Aleppo as in \cite{Xu2019}. First, we obtained imagery from before and after the disaster for each region, mainly from DigitalGlobe’s WorldView 2 and 3 satellites. For Haiti, candid flyover images were provided by the National Oceanic and Atmosphere Administration. We resampled all images to 0.3 meter resolution for consistency. 

Next, we obtained positive ground truth labels from building damage assessments provided by UNOSAT, the operational satellite applications programme of the United Nations Institute for Training and Research (UNITAR), available on the Humanitarian Data Exchange website \cite{hdx}. UNOSAT assessments use a 5-level scale to measure damage, but the labels were noisy and inconsistent across different datasets. Therefore, we grouped “Severe Damage” and “Destroyed” into a single “Damaged” class and formulated our problem as a binary classification problem to identify “Damaged” and “Undamaged” buildings. To acquire Undamaged examples, we used a pretrained building detection model \cite{Xu2019} to identify buildings and filtered out the ones marked as damaged in UNOSAT assessments.

Finally, to create the training examples, we sampled crops centered around each building. We then aligned the pre- and post-disaster imagery and used Google Earth Engine \cite{gorelick2017google} to spatially join the labels and cropped images. Each example in our dataset contains a 6-channel, 64 x 64 image and a classification label (0 for undamaged, 1 for damaged). There are 50,742 Haiti examples (44\% positive), 12,897 Santa Rosa examples (27\% positive), and 10,452 Aleppo examples (44\% positive). In a newly unfolding disaster, this volume of labeled data would not be available. To simulate this, a random sample of the examples were considered labeled and the rest unlabeled (Section \ref{results}).

%% file: sections/methods.tex
We formulate our task as binary classification, where the two classes are undamaged (0) or damaged (1). We define a batch of $B$ labeled examples as $\mathcal{X} = \{(x_b, p_b) : b \in (1, ..., B)\}$, where $x_b$ contains the 6-channel example images and $p_b$ are the one-hot labels. We also define a batch of $\mu B$ unlabeled examples as $\mathcal{U} = \{u_b : 1, .., \mu B\}$, where $\mu$ is a hyperparameter that determines the relative sizes of $\mathcal{X}$ and $\mathcal{U}$. We trained classification models using MixMatch and FixMatch, which use pseudo-labeling to create artificial labels for $\mathcal{U}$. Then, a convolutional neural network (CNN) is trained to predict on augmented examples produced from both $\mathcal{X}$ and $\mathcal{U}$, using a loss function that has a labeled and unlabeled loss term. We compare the differences in pseudo-labeling and loss functions below.

\begin{figure*}
    \centering
    \includegraphics[width=\textwidth]{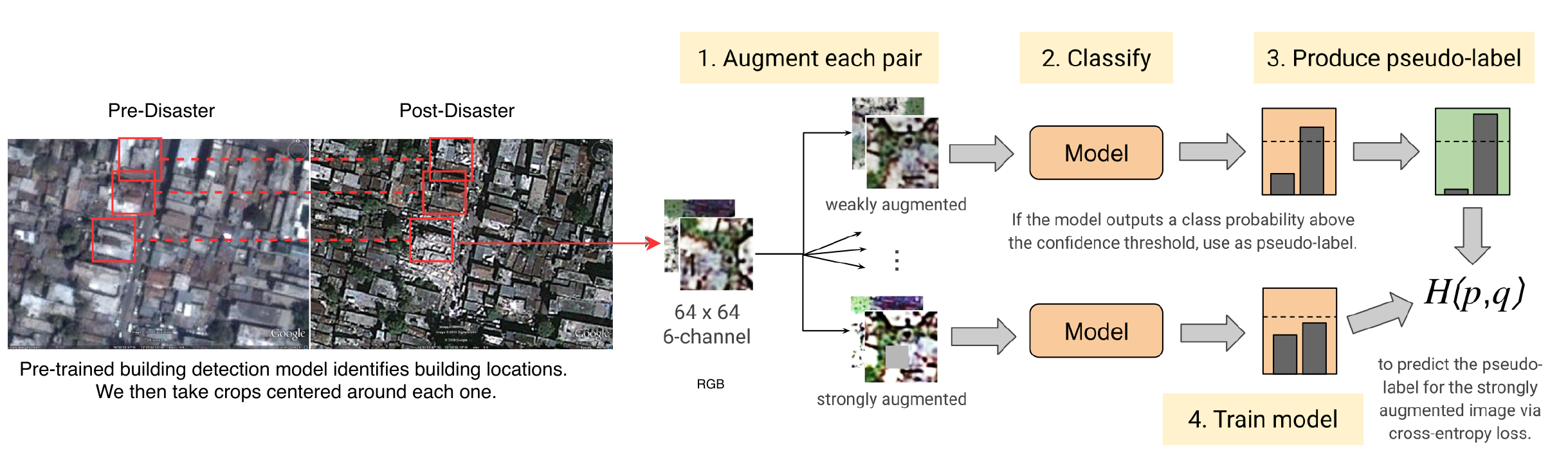}
\caption{Pipeline that shows how the pre- and post-disaster images are stacked into a single 6-channel input and then augmented to produce pseudo-labels for training. This diagram includes strong augmentation to represent FixMatch, but MixMatch uses only weakly augmented images.}
\label{fig:modeldiagram}
\end{figure*}

\paragraph{Producing Pseudo-Labels} We first review the label guessing step of \textbf{MixMatch} \cite{berthelot2019mixmatch}. MixMatch initially applies weak augmentation, which consists of random flips, rotations, and shifts, to both $\mathcal{X}$ and $\mathcal{U}$. Let $\alpha(\cdot)$ denote weak augmentation, such that, for a given unlabeled batch $u_b$, $\alpha(u_b)$ contains $K$ augmentations for each example. We compute the average of the model's predicted class distributions across all augmentations, which is $\bar{q}_b = \frac{1}{K} \sum_{k=1}^{K} p_{model}(y | \alpha(u_{b,k}))$. Then, the method applies sharpening, which reduces the entropy of the label distribution as introduced in \cite{Goodfellow2016}. $q_b = Sharpen(\bar{q}_b)$ acts as the target for the model's prediction on an augmentation of $u_b$. 

Finally, MixMatch takes the augmented versions of the labeled data $\alpha(\mathcal{X}) = ((\alpha(x_b), p_b); b \in (1, ..., B))$ and unlabeled data $\alpha(\mathcal{U}) = ((\alpha(u_b), p_b); b \in (1, ..., B))$ and applies MixUp \cite{zhang2018mixup}, mixing both the labeled and unlabeled examples with label guesses. We refer to the final labeled dataset with augmentations and MixUp applied as $\mathcal{X}'$ and the final unlabeled dataset as $\mathcal{U}'$.

Now, we review the pseudo-labeling step of \textbf{FixMatch} \cite{sohn2020fixmatch}, which presented a relatively simpler method but performed better on benchmark datasets. Similarly to MixMatch, FixMatch creates the targets from the weakly augmented versions of a given unlabeled example. However, without sharpening the output distribution, it uses $\arg\max(q_b)$ as the pseudo-label. FixMatch uses not only weak augmentation but also strong augmentation, denoted by $\mathcal{A}(\cdot)$. In this work, we use FixMatch with RandAugment (RA) \cite{Berthelot2020ReMixMatch}, which randomly applies a series of transformations.
The types of transformations included changing the brightness, contrast, or saturation level, solarizing, posterizing, applying CutOut, and more \cite{sohn2020fixmatch}. FixMatch makes use of both weakly and strongly augmented images by encouraging the model to predict the pseudo-label for the strongly augmented versions of $u_b$, or $\mathcal{A}(u_b)$. We show the implications of this difference by comparing the loss functions below.

\paragraph{Loss Function} For both methods, the loss function has a labeled or supervised loss term, $\mathcal{L}_s$, and an unlabeled loss term, $\mathcal{L}_u$. They are combined into a single training objective as $\mathcal{L} = \mathcal{L}_s + \lambda \mathcal{L}_u$, where $\lambda$ is a hyperparameter that denotes the relative weight of the unlabeled loss. The labeled loss is similar for both methods. In MixMatch, for each batch $x'_b \in \mathcal{X}'$, the labeled loss is $\mathcal{L}_s = \frac{1}{B} \sum_{b=1}^{B} H(p_b, p_{model}(y | x'_b))$, where $H(p,q)$ is defined as the cross-entropy between two probability distributions. In FixMatch, where MixUp is not applied, $\mathcal{L}_s = \frac{1}{B} \sum_{b=1}^{B} H(p_b, p_{model}(y | \alpha(x_b)))$. The unlabeled loss term differs more significantly. For MixMatch, it is the squared $L_2$ loss on predictions and guessed labels as shown by eq. (\ref{mixmatchlu}) for each batch $u'_b \in \mathcal{U}'$.
\begin{equation}
    \mathcal{L}_u = \frac{1}{\mu B} \sum_{b=1}^{B} ||q_b - p_\mathrm{model}(y | u'_b)||_2^2
\label{mixmatchlu}
\end{equation}
Alternatively, FixMatch enforces cross-entropy loss against the strongly augmented unlabeled examples given the weakly augmented ones as shown in eq. (\ref{fixmatchlu}).
\begin{equation}
    \mathcal{L}_u = \frac{1}{\mu B} \sum_{b=1}^{B} \mathbbm{1}(\max(q_b) \geq  \tau) H(\arg\max(q_b), p_\mathrm{model}(y | \mathcal{A}(u_b)))
\label{fixmatchlu}    
\end{equation}
where $\tau$ is a scalar hyperparameter that signifies the threshold
above which we retain a pseudo-label. Although Fixmatch's approach is simpler and empirically performed better, we ran experiments with both methods because of the shift in domain to satellite imagery, where augmentations may have different semantic implications. For example, applying a weak augmentation of a random shift may cause the model to predict a building as damaged because buildings  can shift in an earthquake. Alternatively, due to the noisy nature of satellite imagery, strong augmentations could make it more challenging for the model to learn. Thus, we tried both MixMatch and FixMatch in our experiments.


%% file: sections/results.tex
We test the efficacy of SSL methods on classifying buildings in three different regions: Santa Rosa~\cite{gupta2019xbd}, Haiti, and Aleppo~\cite{Xu2019}, using four different methods: the Twin Tower model that performed best in the past work by Xu et al. \cite{Xu2019} as a baseline,  fully-supervised learning with only labeled training data, and semi-supervised learning with MixMatch~\cite{berthelot2019mixmatch} and FixMatch~\cite{sohn2020fixmatch}.

\textbf{Setting.~} We split datasets into 90\% and 10\% for train and test, respectively. For SSL experiments, we randomly sample a specific number of examples (e.g., 10, 50, 100, 500) evenly from each class as the labeled training set and consider the remainder as the unlabeled set. We conduct experiments $5$ times with different splits of the labeled training set.
We train a variant of Wide Residual Network (WRN)~\cite{zagoruyko2016wide}, containing four residual blocks with 32, 64, 128, and 256 filters, respectively. We closely follow the training strategies in~\cite{sohn2020fixmatch}, using momentum (0.9) SGD with cosine learning rate decay. We tune other hyperparameters, such as the learning rate or weight decay, for each method, but they are shared across datasets. We provide experimental details in the supplementary material.


\textbf{Results.~} In Table~\ref{tab:single_region_results}, we report accuracy averaged over 5 runs, each of which is trained on different labeled data splits. We also train a fully supervised model on all labeled data (called "90\% supervised") as a performance upper bound, although this would not be available in practice. 

SSL improves accuracy on all three disaster datasets. MixMatch shows consistent improvements over Twin Tower and fully-supervised models in Haiti and Aleppo, regardless of the number of labeled training data. FixMatch makes further significant improvements in all three regions, although it does require 100 or more labeled data to surpass MixMatch performance in Haiti. In Aleppo, FixMatch with $500$ labeled data is able to outperform the fully-supervised model trained on all labeled data, possibly because strong augmentations capture a level of variance not in the dataset alone. We examine augmentation policies in Appendix \ref{sec:ablation_augmentation_policy}.
Overall, our results demonstrate the generality of modern SSL methods based on data augmentation, such as MixUp~\cite{zhang2018mixup}, CTAugment~\cite{Berthelot2020ReMixMatch}, or RandAugment~\cite{cubuk2019randaugment}, in spite of the contrasting image statistics of satellite imagery compared to standard visual recognition benchmarks, such as CIFAR-10~\cite{krizhevsky2009learning} or ImageNet~\cite{deng2009imagenet}. 

%


\begin{table}[t]
    \centering
    \small{
    \begin{tabular}{l|c|cccc}
        \toprule
        \textbf{Dataset} &
        \textbf{\# Labeled Data} &
        \textbf{Twin Tower} &
        \textbf{Fully Supervised} & \textbf{MixMatch} &  \textbf{FixMatch} \\ 
        \midrule
        \multirow{5}{*}{Haiti} & 10 &
        0.53{\scriptsize$\pm0.03$} &
        0.58{\scriptsize$\pm0.02$} & 0.60{\scriptsize$\pm0.11$} & 0.56{\scriptsize$\pm0.05$} \\
        & 50 &
        0.56{\scriptsize$\pm0.02$} &
        0.64{\scriptsize$\pm0.01$} & 0.72{\scriptsize$\pm0.03$} & 0.61{\scriptsize$\pm0.03$} \\ 
        & 100 &
        0.56{\scriptsize$\pm0.05$} &
        0.69{\scriptsize$\pm0.02$} & 0.75{\scriptsize$\pm0.04$} & 0.75{\scriptsize$\pm0.10$} \\
        & 500 &
        0.71{\scriptsize$\pm0.01$} &
        0.75{\scriptsize$\pm0.01$} & 0.82{\scriptsize$\pm0.01$} & \textbf{0.87}{\scriptsize$\pm0.01$} \\
        & 90\% supervised (45,667 data) & 0.90 & - & -  \\
        \midrule
        \multirow{5}{*}{Santa Rosa} & 10 & 
        0.54{\scriptsize$\pm0.03$} &
        0.74{\scriptsize$\pm0.08$} & 0.70{\scriptsize$\pm0.06$} & 0.92{\scriptsize$\pm0.07$} \\
        & 50 &
        0.54{\scriptsize$\pm0.03$} &
        0.91{\scriptsize$\pm0.03$} & 0.85{\scriptsize$\pm0.08$} & 0.96{\scriptsize$\pm0.02$} \\ 
        & 100 &
        0.58{\scriptsize$\pm0.04$} &
        0.92{\scriptsize$\pm0.01$} & 0.88{\scriptsize$\pm0.04$} & 0.97{\scriptsize$\pm0.01$} \\
        & 500 &
        0.69{\scriptsize$\pm0.03$} &
        0.96{\scriptsize$\pm0.01$} & 0.96{\scriptsize$\pm0.01$} & \textbf{0.98}{\scriptsize$\pm0.00$} \\
        & 90\% supervised (11,067 data) & 0.99 & - & - \\
        \midrule
        \multirow{5}{*}{Aleppo} & 10 &
        0.52{\scriptsize$\pm0.04$} &
        0.55{\scriptsize$\pm0.05$} & 0.65{\scriptsize$\pm0.08$} & 0.72{\scriptsize$\pm0.15$} \\
        & 50 &
        0.56{\scriptsize$\pm0.02$} &
        0.65{\scriptsize$\pm0.06$} & 0.73{\scriptsize$\pm0.06$} & 0.88{\scriptsize$\pm0.01$} \\ 
        & 100 &
        0.57{\scriptsize$\pm0.04$} &
        0.71{\scriptsize$\pm0.06$} & 0.78{\scriptsize$\pm0.04$} & 0.89{\scriptsize$\pm0.01$} \\
        & 500 &
        0.67{\scriptsize$\pm0.04$} &
        0.82{\scriptsize$\pm0.01$} & 0.85{\scriptsize$\pm0.02$} & \textbf{0.90}{\scriptsize$\pm0.01$} \\
        & 90\% supervised (9,406 data) & 0.88 & - & -  \\
        \bottomrule
    \end{tabular}
    }
    \vspace{0.05in}
    \caption{
    Classification accuracy of the Twin Tower~\cite{Xu2019} baseline, fully-supervised, MixMatch, and FixMatch models with varying amounts of labeled training data, averaged over 5 runs (each with a different labeled data split). To provide an upper bound, one model is trained using all training data.}
    \label{tab:single_region_results}
\end{table}

\begin{figure*}
    \centering
    \includegraphics[width=\textwidth]{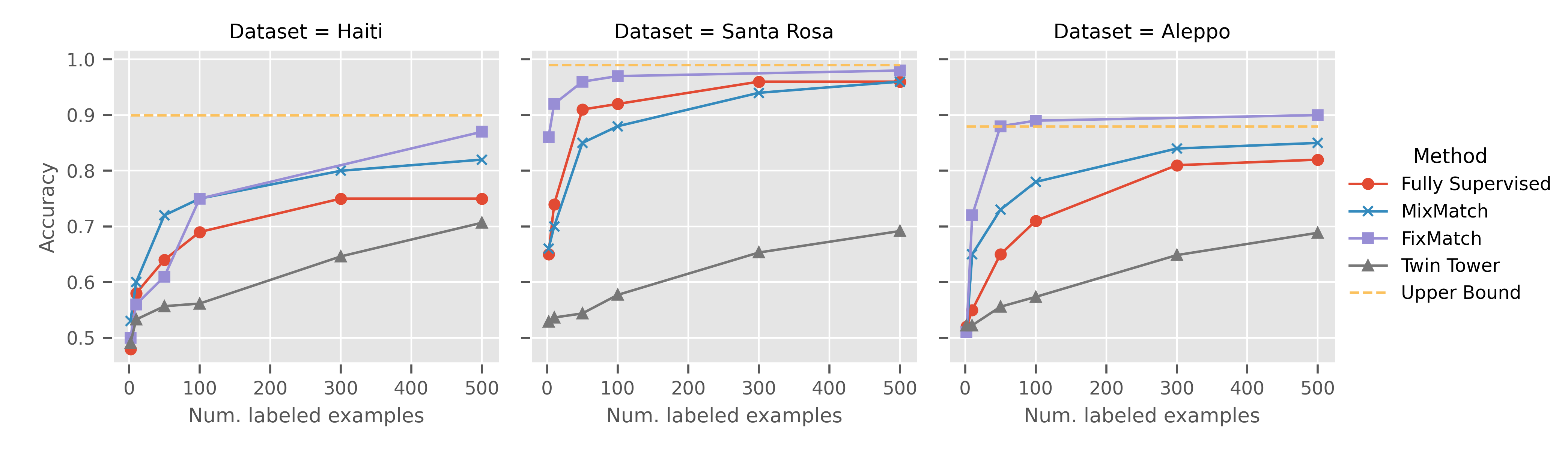}
\caption{Performance of fully and semi-supervised models by number of labeled training examples.}
\label{fig:resultsgraph}
\end{figure*}

%% file: sections/conclusion.tex
In this paper, we introduced a novel application of semi-supervised learning to automatically detect damaged buildings in satellite imagery with limited labeled data. We experimented with two recent techniques, MixMatch and FixMatch, and showed how they are able to achieve strong performance 100 labeled examples or fewer by leveraging unlabeled data. They consistently outperformed fully supervised models and even achieved performance close to that of a fully supervised setting with no data constraints. The results empirically showed how SSL approaches can be useful to train models when a new disaster is unfolding in an unseen region. For future work, we plan to investigate how to effectively incorporate data from past disasters; there may be region-independent transformations caused by a disaster that the models do not sufficiently capture or different types of augmentations and losses that are more robust to the noise inherent to satellite imagery. 
